\pdfoutput=1

\documentclass[11pt]{article}

\PassOptionsToPackage{table}{xcolor}
\usepackage[final]{acl}

\usepackage{times}
\usepackage{latexsym}
\usepackage[T1]{fontenc}
\usepackage[utf8]{inputenc}
\usepackage{microtype}
\usepackage{inconsolata}
\usepackage{graphicx}
\usepackage{amsmath}
\usepackage{enumitem}
\usepackage{graphicx}    
\usepackage{subcaption}  
\usepackage{caption}
\usepackage[utf8]{inputenc}
\usepackage{CJKutf8}
\usepackage{wrapfig} 
\usepackage{pifont}
\usepackage{float}
\usepackage{booktabs}
\usepackage[most]{tcolorbox}
\usepackage{multirow}

\title{The Impact of Language Mixing on Bilingual LLM Reasoning}

\author{%
  \textbf{Yihao Li\textsuperscript{1}},
  \textbf{Jiayi Xin\textsuperscript{1}},
  \textbf{Miranda Muqing Miao\textsuperscript{1}},
  \textbf{Qi Long\textsuperscript{1}},
  \textbf{Lyle Ungar\textsuperscript{1}}\\[1ex]
  \textsuperscript{1}University of Pennsylvania \\
  \texttt{\{liyihao, jiayixin, miaom\}@seas.upenn.edu} \\
  \texttt{qlong@upenn.edu, ungar@cis.upenn.edu}
}

\begin{document}
\maketitle
\begin{abstract}

Proficient multilingual speakers often intentionally switch languages in the middle of a conversation. Similarly, recent reasoning-focused bilingual large language models (LLMs) with strong capabilities in both languages exhibit \textbf{language mixing}—alternating languages within their chain of thought. Discouraging this behavior in DeepSeek-R1 was found to degrade accuracy, suggesting that language mixing may benefit reasoning. In this work, we study language switching in Chinese-English bilingual reasoning models. We identify reinforcement learning with verifiable rewards (RLVR) as the critical training stage that leads to language mixing. We show that language mixing can enhance reasoning: enforcing monolingual decoding reduces accuracy by 5.6 percentage points on \texttt{MATH500}. Additionally, a lightweight probe can be trained to predict whether a potential language switch would benefit or harm reasoning, and when used to guide decoding, increases accuracy by 2.92 percentage points. Our findings suggest that language mixing is not merely a byproduct of multilingual training, but is a \textit{strategic reasoning behavior}.\footnote{Code available at \url{https://github.com/liyihao0302/language-mixing-in-llm-reasoning}}

\end{abstract}

\section{Introduction}
Multilingual speakers sometimes mix languages during reasoning, which is a phenomenon known in linguistics as \textit{code-switching} \cite{appel2005language, ozkara2025code}. Though switching languages seems to disrupt coherence, multilingual speakers persist in this behavior for practical reasons. Languages vary in how they organize thoughts and some express certain concepts (e.g., numbers) more efficiently than others \citep{boroditsky2001does, haun2011plasticity, pica2004exact, miura1988effects}. Language mixing helps them express ideas more precisely, fill lexical gaps when one language falls short \cite{kuzyk2020there}, and reduce cognitive load by directing more mental effort toward the reasoning task itself \cite{lehti2013code}.

Large Language Models (LLMs) have evolved from English-centric models to those with strong multilingual abilities, with some achieving true bilingualism through balanced English-Chinese training~\cite{liu2024deepseek, qwen2025qwen25technicalreport}. How these bilingual models differ from primarily monolingual LLMs raises intriguing questions for computational linguists. One striking phenomenon in this space is language mixing, with recent state-of-the-art RL-trained English-Chinese bilingual LLMs such as DeepSeek‑R1~\cite{guo2025deepseek} and QwQ‑32B~\cite{qwen2024qwqpreview} displaying human-like language mixing behavior in their chain-of-thought: they respond in languages different from the prompt and switch languages (sometimes repeatedly) during their reasoning process.

\definecolor{mypurple}{RGB}{128,0,128}
\definecolor{darkgreen}{RGB}{0,100,0}

\begin{figure}[!htbp]  
    \centering
    \includegraphics[width=0.49\textwidth]{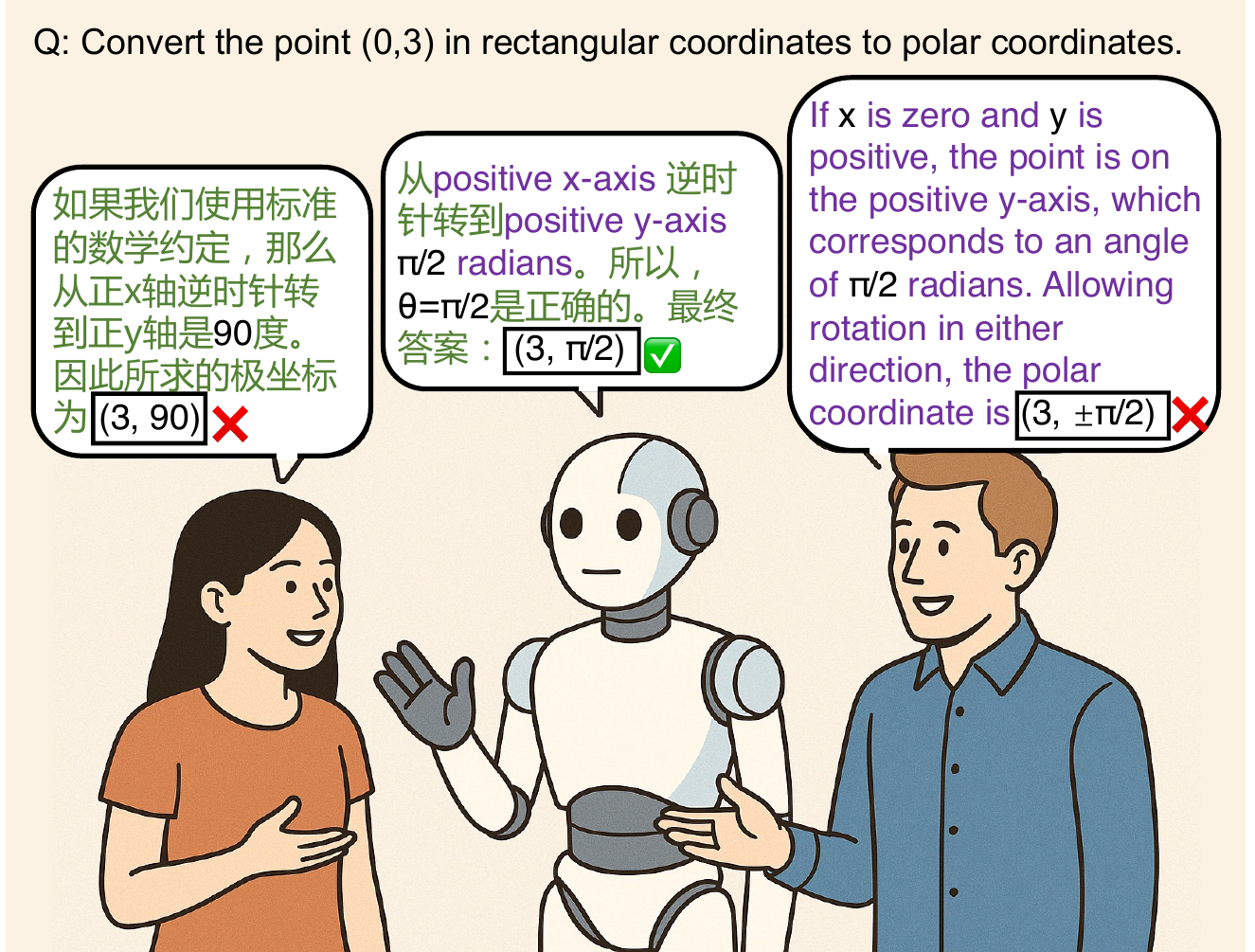}  
    \caption{An illustration of bilingual code‑switching improving reasoning performance. Two monolingual speakers, one in \textcolor{darkgreen}{Chinese} and the other in \textcolor{mypurple}{English}, fail to solve a math problem, while an LLM robot that code‑switches between both succeeds. Black text denotes language‑agnostic content.}
    \vspace{-0.5em}
    \label{fig:code_switch}
\end{figure}

\begin{figure*}[!htbp]  
    \centering
    \includegraphics[width=0.98\textwidth]{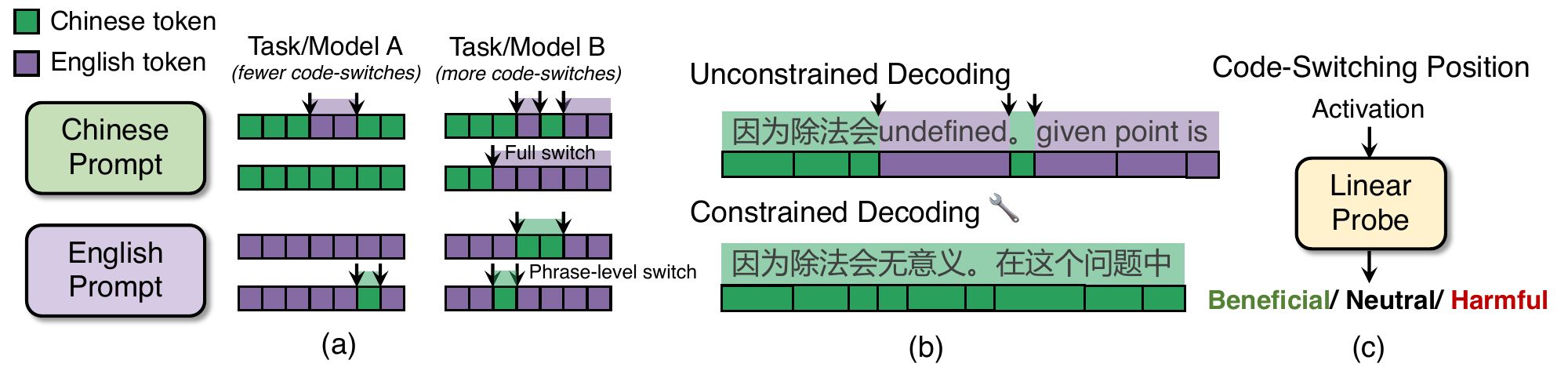}  
    \caption{Overview of our analysis of language mixing in LLM reasoning. (a) We identify common language mixing patterns and triggers that lead to increased language mixing  (Section~\ref{sec:patterns}). (b) We compare unconstrained bilingual outputs with constrained monolingual outputs to evaluate the impact of language mixing on reasoning performance  (Section~\ref{sec:sec3}). (c) We train a probe to classify code-switches as \{\texttt{\textcolor{darkgreen}{Beneficial}}, \texttt{Neutral}, or \texttt{\textcolor{red}{Harmful}}\}, and use it to guide decoding  (Section~\ref{sec:probe}).
}
    \label{fig:overview}  
\end{figure*}
Proficient multilingual speakers of both languages can benefit from reasoning with code-switching. Can LLMs similarly benefit? (Figure~\ref{fig:code_switch}). The parallel seems plausible: both humans and LLMs potentially share needs for expressivity, precision, reduced cognitive load, and efficiency (which for LLMs translates to using fewer tokens and shorter context windows). Supporting this, DeepSeek-R1 demonstrates a performance degradation when a language consistency reward is introduced during training~\cite{guo2025deepseek}. These findings motivate our study into how language mixing affects LLM reasoning, centered on a key question: \textit{Do LLMs reason better or worse with English-Chinese language mixing?}


To study whether language mixing \textbf{causally} improves reasoning performance, we first observe that stronger reasoning \textbf{correlates} with increased mixing, and identify reinforcement learning with verifiable rewards (RLVR)~\cite{lambert2024tulu} as the critical training stage that induces it (Section 2). We then test \textbf{causation} through interventions in both directions: a) \textit{decreasing} language mixing through constrained monolingual decoding degrades performance (Section 3); b) \textit{strategically enhancing} language mixing through probe-guided decoding improves performance (Section 4). Together, these findings suggest language mixing is not a random artifact but a potentially deliberate, \textit{useful strategy} for enhancing LLM reasoning.

Our contributions are summarized as follows:
\begin{itemize}[leftmargin=*, itemsep=0pt, topsep=0pt]
    \item[$\star$] We demonstrate that bilingual chain‑of‑thought reasoning with language mixing \textbf{causally} enhances performance: unconstrained bilingual outputs significantly outperforming monolingually constrained ones (p<0.05), and probe‑guided yields further gains.
    \item[$\star$] We identify RLVR as the critical training stage that triggers language mixing, suggesting this behavior may emerge from natural optimization.
    \item[$\star$] We introduce probe‑guided decoding, which integrates a lightweight real‑time probe in the generation loop to trigger beneficial bilingual switches, improving reasoning with minimal overhead.
\end{itemize}

\section{Where does Language Mixing Occur?}\label{sec:patterns}
\subsection{Detecting Code-Switches}

\definecolor{lightgray}{gray}{0.9}
\definecolor{lightpeach}{RGB}{255,228,225}
\begin{table*}[t]
\centering
\caption{Language‐mixing statistics across QwQ and DeepSeek‑R1 series for Chinese (ZH) and English (EN) prompts.  
\textbf{\%Prob.}: percentage of problems with code-switch;
\textbf{Switch}: average number of switches per problem;
\textbf{Tokens/Switch}: mean tokens between consecutive switches;
\textbf{Non-prompt (\%)}: fraction of tokens in a language different from the prompt;
\textbf{White}: pretraining only; \textbf{Grey}: SFT + RLHF; \textbf{Pink}: RLVR. DeepSeek-V3-Base is excluded from interpretation due to frequent generation of irrelevant content.}
\label{tab:langmix-stats}
\resizebox{\textwidth}{!}{%
\begin{tabular}{l
    rrrr
    rrrr
}
\toprule
& \multicolumn{4}{c}{\textbf{ZH}} & \multicolumn{4}{c}{\textbf{EN}} \\
\cmidrule(lr){2-5} \cmidrule(lr){6-9}
\textbf{Model}
& \% Prob. & Switch & Tokens/Switch & Non-prompt (\%)
& \% Prob. & Switch & Tokens/Switch & Non-prompt (\%) \\
\midrule
\rowcolor{white}
Qwen2.5‑32B
  & 14.8\%  &  1.98  &   667.96   & 1.42\%
  & 0.0\%   & 0.00   & 0.00               & 0.00\% \\

\rowcolor{lightgray}
Qwen2.5‑32B‑Instruct
  & 8.8\%   &  0.36  &  1986.71   & 0.23\%
  & 0.0\%   & 0.00   & 0.00               & 0.00\% \\

\rowcolor{lightpeach}
QwQ32B‑Preview
  & \textbf{77.4\%}  &  \textbf{7.22}  &   \textbf{217.03}   & \textbf{4.28\%}
  & \textbf{0.6\%}   & \textbf{0.02}  & $\mathbf{1.50 \times 10^{5}}$
& \textbf{0.00\%} \\

\rowcolor{lightpeach}
QwQ32B
  & 29.2\%  &  6.20  &   585.85   & 0.48\%
  & 0.5\%   & 0.01   & $2.85\times10^{5}$ & 0.00\% \\
\midrule
\rowcolor{white}
DeepSeek‑V3‑Base
  & 32.2\%  &  9.95  &   190.78   & 2.53\%
  & 4.2\%   & 1.51   &   980.76           & 1.18\% \\

\rowcolor{lightgray}
DeepSeek‑V3
  & 8.4\%   &  0.39  &  3574.98   & 0.08\%
  & \textbf{0.4\%}   & \textbf{0.01}   & $\mathbf{1.50\times10^{5}}$
 & \textbf{0.02\%} \\

\rowcolor{lightpeach}
DeepSeek‑R1‑Zero
  & 10.9\%  &  0.21  &  7048.94   & 0.82\%
  & 0.0\%   & 0.00   & 0.00               & 0.00\% \\

\rowcolor{lightpeach}
DeepSeek‑R1
  & \textbf{27.1\%}  &  \textbf{4.39}  &   \textbf{688.31}   & \textbf{0.38\%}
  & 0.0\%   & 0.00   & 0.00               & 0.00\% \\

\rowcolor{lightpeach}
DeepSeek‑R1‑Distill‑Llama‑8B
  & 23.6\%  &  2.46  &  1128.53   & 0.31\%
  & 0.0\%   & 0.00   & 0.00               & 0.00\% \\

\rowcolor{lightpeach}
DeepSeek‑R1‑Distill‑Qwen‑32B
  & 21.2\%  &  1.94  &  1292.15   & 0.24\%
  & 0.0\%   & 0.00   & 0.00               & 0.00\% \\
\bottomrule
\end{tabular}%
}
\end{table*}
Code-switching, by definition, means switching between languages in a single conversation. As illustrated in Fig.\ref{fig:overview}(a), segments of Chinese (in green) and segments of English (in purple) alternate, and these transitions represent code-switching occurrences. In written text, elements such as mathematical expressions or code (typically composed of English tokens) are language-agnostic and universally used across speakers of different languages. Thus, a paragraph written in Chinese that includes mathematical expressions using English tokens should not be considered language mixing. We define a \textit{code-switching position} as the first \textbf{text} token (in either English or Chinese) where the language switches from one to another, excluding any language-agnostic content such as math expressions. These positions correspond to the arrow markers shown in Fig.~\ref{fig:overview}(a).

Based on this definition, we implement a rule-based procedure to detect Chinese-English code-switching. We first filter out language-agnostic tokens (e.g., LaTeX math) and then distinguish English from Chinese tokens using ASCII encoding (see Appendix~\ref{sec:appendix_detection} for details). We evaluate code‑switching behavior with three key statistics on bilingual datasets that contain parallel English–Chinese versions of each problem (by translating from the original language):
\begin{itemize}
\item \textbf{Switch count:} The total number of switches (back and forth) between languages when processing problems under English and Chinese prompts.
\item \textbf{Tokens between switches:} The average number of tokens generated between consecutive language switches, quantifying how frequently the model alternates between languages measured in tokens.
\item \textbf{Non-prompt language fraction:} The fraction of tokens generated in a language different from the prompt language (shown as the shaded area between arrows in Fig.~\ref{fig:overview}(a)), measuring how long the model stays in the non-prompt language. This evaluates the extent and persistence of language mixing.
\end{itemize}

\subsection{Tracing the Evolution of Language Mixing in LLMs}

The development of multilingual reasoning LLMs has followed a temporal progression of introducing training strategies, with the introduction of generative pre-training, post-training methods such as supervised fine-tuning (SFT), reinforcement learning with human feedback (RLHF)~\cite{ouyang2022training}, and most recently, reinforcement learning with verifiable rewards (RLVR). RLVR replaces learned reward models with simple "pass/fail" checks that assign fixed rewards for correct outcome~\cite{lambert2024tulu}, achieving remarkable gains in reasoning~\cite{chen2025sft, xie2025logic}. Frequent English–Chinese mixing between the two highest‑resource languages in open‑weight models like QwQ32B~\cite{qwen2024qwqpreview} and DeepSeek‑R1~\cite{guo2025deepseek} is a recent observation, likely triggered by the newly popularized RLVR training strategy. Here, we aim to identify exactly which training stage(s) trigger such pronounced English–Chinese mixing.

\begin{figure*}[!htbp]  
    \centering
    \includegraphics[width=1.0\textwidth]{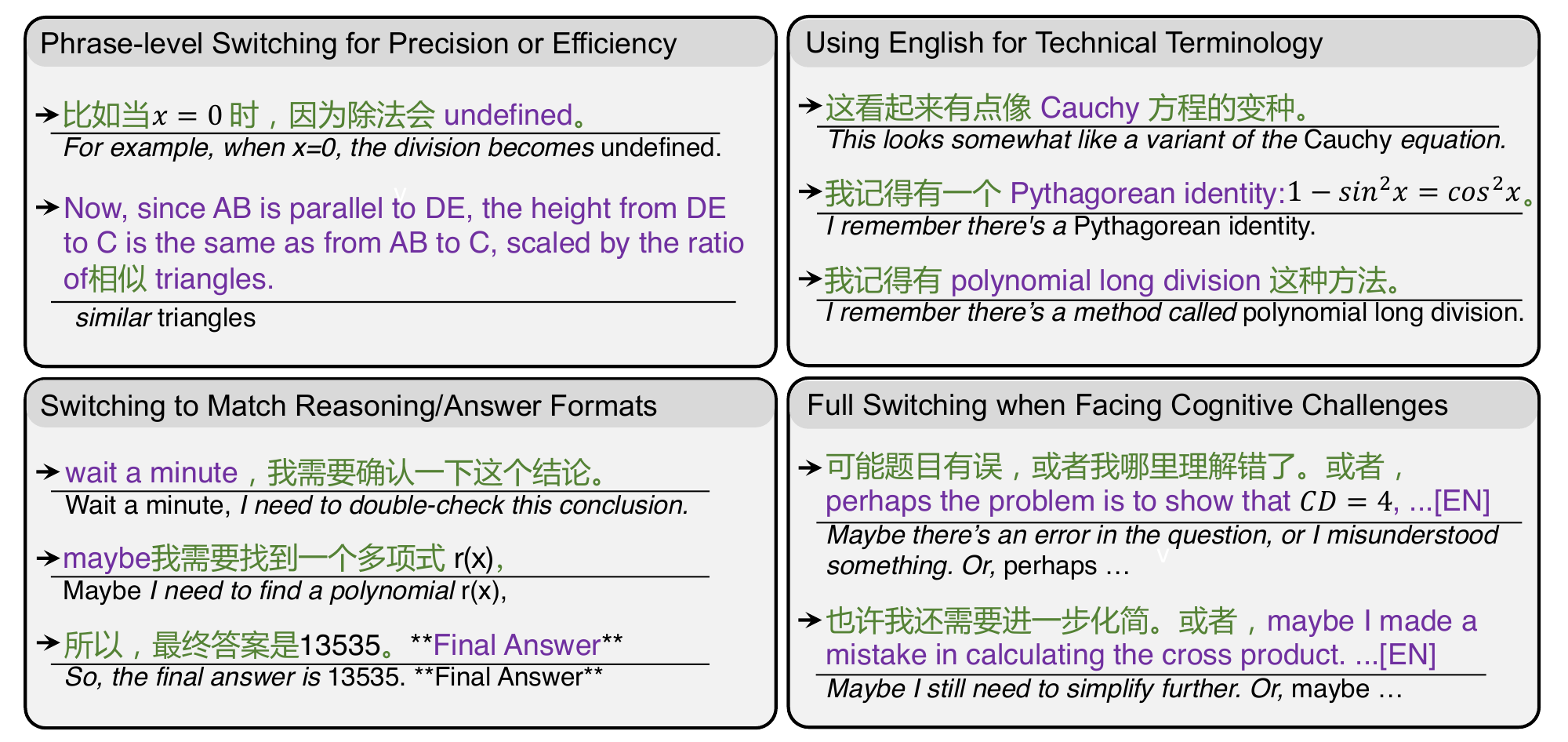}  
    \caption{Four patterns of code-switching observed in LLM outputs. \textbf{Top left:} Phrase-level switching, often short and used for precision or efficiency. \textbf{Top Right:} Switching to English for technical terms. \textbf{Bottom left:} Switching to match reasoning or answer formats. \textbf{Bottom right:} Full switch to another language when the model is unable to find a solution.}
    \label{fig:patterns}  
\end{figure*}

To do so, we trace the evolution of language-mixing behavior across iterations of QwQ32B and DeepSeek-R1 models. For the QwQ series, we examine Qwen2.5-32B (base model with pre-training only), Qwen2.5-32B-instruct (post-trained with SFT and RLHF)~\cite{qwen2025qwen25technicalreport}, and two generations trained with RLVR: QwQ32B-preview~\cite{qwen2024qwqpreview} and QwQ32B~\cite{qwen2025qwq32b}. For the DeepSeek-R1 series, we analyze DeepSeek-V3-base (the foundation model with only pretraining), DeepSeek-V3 (with SFT and RLHF applied)~\cite{liu2024deepseek}, DeepSeek-R1-zero (a version without language consistency reward, where language mixing was documented), DeepSeek-R1 (with language consistency reward implemented), and various DeepSeek-R1 distilled variants~\cite{guo2025deepseek}. We evaluate language mixing occurrences across these models using MATH500 (in both English and Chinese versions). To ensure comparable analysis, we prompt base and instruct models for lengthy chain-of-thought reasoning to match output lengths (see Appendix~\ref{sec:appendix_rlvr}).

Comparing the three model groups in Table~\ref{tab:langmix-stats}—base (white), SFT/RLHF (grey), and RLVR (pink)—we observe that RLVR models exhibit the most language mixing, followed by pre-trained base models, while SFT/RLHF models demonstrate the least mixing behavior.

\paragraph{Pretraining.}During large-scale pretraining, LLMs are exposed to web-scale multilingual corpora, yet they rarely encounter natural code-switched input, which is far more common in speech than in text. But training data contains natural code‑switching~\cite{wu2025enhancing}, so LLMs may learn to code‑switch to some extent. As Table~\ref{tab:langmix-stats} shows (white rows), Qwen2.5-32B exhibits minimal but non-zero code-switching, while DeepSeek-V3-Base displays frequent switches. However, we discount DeepSeek-V3-Base's behavior as it tends to ramble with irrelevant content without proper ability to terminate generation with [EOS].

\paragraph{SFT and RLHF.}Supervised fine‑tuning (SFT) trains on human curated, high‑quality responses that are predominantly monolingual for human readability, and reinforcement learning from human feedback (RLHF) \cite{ouyang2022training} reinforces these monolingual outputs by assigning them higher rewards. As a result, language mixing is strongly suppressed: Qwen2.5‑32B‑Instruct and DeepSeek‑V3‑Instruct exhibit the fewest code‑switching instances among the model families.

\begin{figure*}[!t]
  \centering
  \begin{minipage}[t]{1\textwidth}
    \centering
    \includegraphics[width=\textwidth]{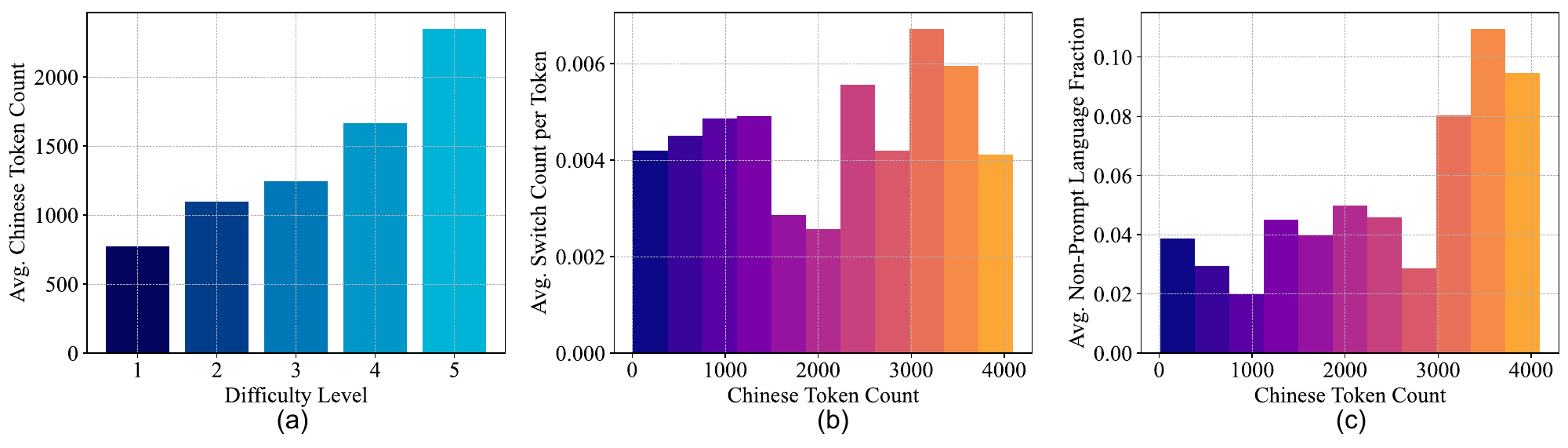}
  \end{minipage}%
  \caption{Quantitative analysis of language-mixing behavior in \texttt{Math500} responses. (a) Correlation between problem difficulty level and response token count for Chinese prompts. (b) Normalized switch count and non-prompt language fraction as functions of token count, showing both code-switching frequency and non-prompt language use increase as chain-of-thought reasoning lengthens.}
  \label{fig:all_plots}
\end{figure*}

\paragraph{RLVR.}RLVR does not constrain the reasoning chain—what matters most is reaching the correct outcome. By optimizing directly for outcome-based rewards, it explores a much larger search space without relying on human priors (similar to the shift from AlphaGo to AlphaZero~\cite{silver2017masteringchessshogiselfplay}). We observe that RLVR models produce frequent code-switching in both families, with QwQ-32B-preview exhibiting the most mixing at 77.4\% for Chinese problems and 0.6\% for English problems. As we later show, trajectories that mix languages tend to achieve higher accuracy, suggesting that RLVR’s outcome‑focused optimization naturally encourages language mixing when it leads to better performance.

To understand why RLVR triggers language mixing, we analyze checkpoints from multiple training steps using Tina-Open-RS1~\cite{wang2025tinatinyreasoningmodels}. Our findings reveal that language mixing increases progressively with RL training steps, with code-switching responses consistently outperforming monolingual responses within trajectory groups. GRPO consequently upweights these higher-advantage code-switching responses, reinforcing the mixing behavior (detailed experimental settings and results in Appendix~\ref{sec:appendix_rlvr}).

Finally, we note findings within model families. Contrary to claims in their paper~\cite{guo2025deepseek}, DeepSeek-R1-Zero displays fewer code-switching instances than DeepSeek-R1 in our testing. This discrepancy may result from our use of greedy decoding. QwQ32B-preview exhibits substantially more code-switching than the newer QwQ32B release, potentially hinting at added language consistency constraints in the updated model. We select QwQ32B-preview for our subsequent analyses.

\subsection{Characterizing Code-Switching Behavior}

\begin{CJK*}{UTF8}{gbsn}
\paragraph{Code-switching patterns.}Based on analysis of QwQ32B-Preview outputs, we identify four main patterns of switches as shown in Figure \ref{fig:patterns}. The most common pattern is phrase-level switching in the top-left examples of Figure~\ref{fig:patterns}, driven primarily by a need for precision or efficiency. Certain concepts may be more clearly expressed in one language, with less ambiguity and often using fewer tokens. For instance, the use of \textit{undefined} in the first example is more precise and less ambiguous than its Chinese counterparts: \textit{无意义} (which can mean “meaningless,” as in “He felt his effort was meaningless”) or \textit{未定义} (which may imply something is not yet defined but could be). It also requires fewer tokens—\textit{undefined} is a single token, while both Chinese alternatives require two.

\end{CJK*}

The second pattern (top right) involves switching to English for domain-specific terminology, likely because the model has limited capacity to store specialized translations across multiple languages. The third pattern (bottom left) shows language switching to conform to specific reasoning or answer formats, such as interjecting "wait, let me double check this" or concluding with "Final answer: ..." in English within otherwise Chinese responses. These formats may originate from supervised fine-tuning on data containing such patterns or reflect the model's emergent self-reflective cues that aren't well-aligned across languages. The fourth pattern (bottom right) involves switching entirely to another language when the model encounters difficulties or recognizes errors in its reasoning. This behavior may suggest a strategy to "clear its mind" or to seek cues in another language. However, this pattern typically appears in more challenging problems, and even after switching languages, the model often fails to reach the correct solution.

These patterns, particularly the first two, mirror common human multilingual behavior and reveal the specific mechanisms through which language mixing may enhance reasoning.

\paragraph{Quantifying language mixing behavior.} In QwQ-32B-preview responses to the \texttt{MATH500} dataset, 77.4\% of answers to Chinese prompts exhibit language mixing, with an average of 7.22 code-switches per problem, compared to just 0.6\% for English prompts. It is already notable that English prompts (with math expressions fully in English tokens) occasionally trigger Chinese token generation. However, Chinese-to-English switching occurs far more frequently, indicating that English remains the model’s dominant or preferred language for reasoning.

We analyze how language mixing behavior relates to problem complexity and response length. Figure~\ref{fig:all_plots}(a) demonstrates the correlation between token count in responses to Chinese prompts and \texttt{MATH500} problem difficulty levels (5 discrete levels). Figures~\ref{fig:all_plots}(b) and (c) quantify switch frequency and non-prompt language fraction as functions of token count. Since these statistics are normalized by token count, we can conclude that longer chain-of-thought reasoning exhibits slightly increased code-switching frequency and a growing fraction of non-prompt language use. This indicates that when tackling more difficult problems, the model adopts as a strategy to use greater language mixing—both switching between languages more frequently and shifting more toward the non-prompt language.

\begin{figure}[!htbp]  
    \centering
    \vspace{-1em}
    \includegraphics[width=0.48\textwidth]{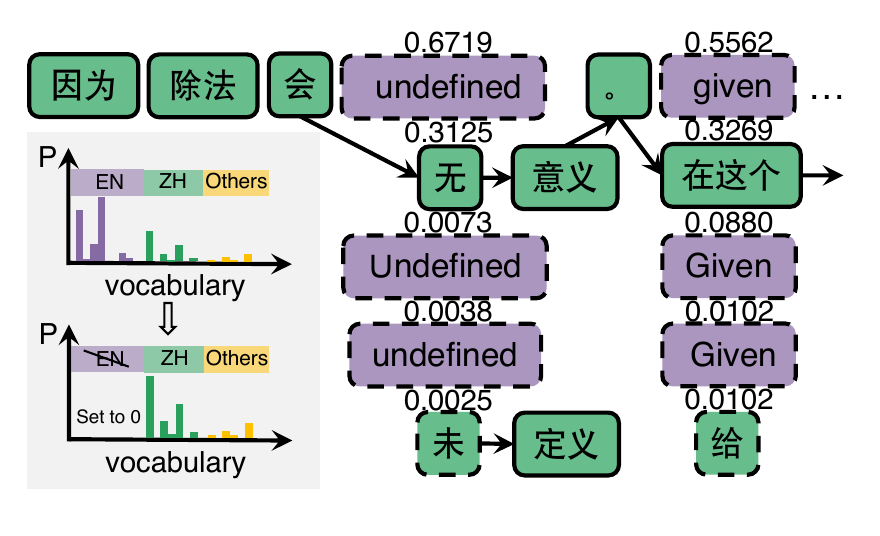}  
    \caption{Token-level constrained decoding: We mask out tokens from the undesired language, forcing generation in the target language.}
    \vspace{-1em}
    \label{fig:constrained_decoding}  
\end{figure}

\section{Do LLMs reason better or worse with language mixing?}\label{sec:sec3}

\subsection{Constrained Decoding}
\begin{figure*}[!t]
  \centering
  \begin{minipage}[t]{1\textwidth}
    \centering
    \includegraphics[width=\textwidth]{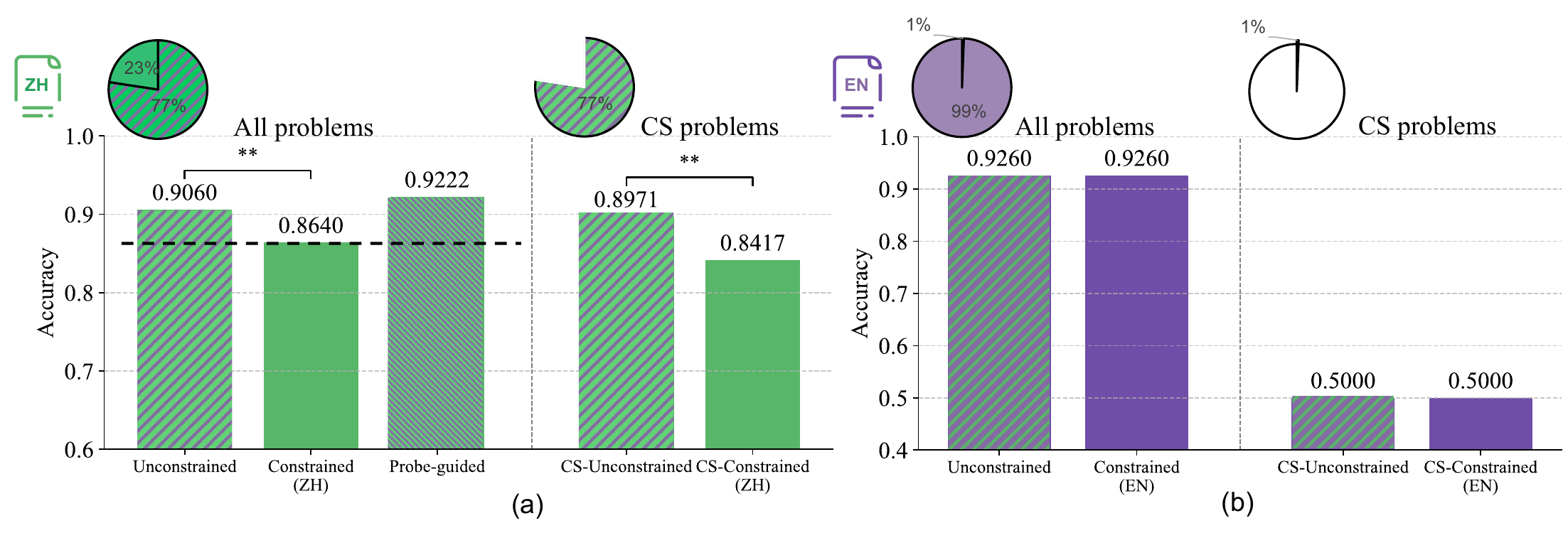}
  \end{minipage}%
  \caption{Comparison of accuracies for unconstrained, constrained, and probe‑guided decoding on \texttt{MATH500}. \textbf{(a)} Chinese prompts (green); \textbf{(b)} English prompts (purple). Bar charts show accuracy for Chinese‑only (solid green), English‑only (solid purple), and language‑mixing (hatched) outputs. Pie charts indicate the proportion of code‑switching (CS) problems. Under Chinese prompts, unconstrained decoding significantly outperforms constrained monolingual decoding.}
  \label{fig:bar1}
\end{figure*}

\begin{figure*}[!t]
  \centering
  \begin{minipage}[t]{1\textwidth}
    \centering
    \includegraphics[width=\textwidth]{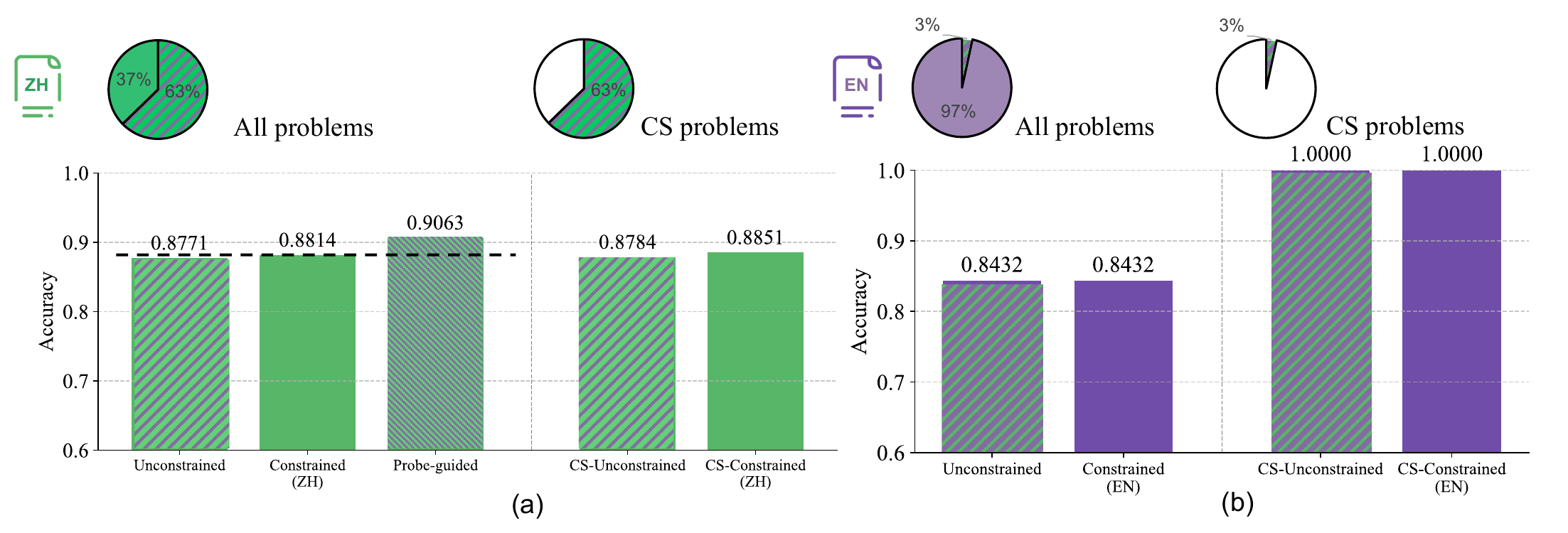}
  \end{minipage}%
  \caption{Comparison of accuracies for unconstrained, constrained, and probe‑guided decoding on \texttt{Gaokao Cloze}. \textbf{(a)} Chinese prompts (green); \textbf{(b)} English prompts (purple).}
  \label{fig:bar2}
\end{figure*}

We expect that reasoning trajectories differ between monolingual and bilingual thinking, as languages have different structural focuses and are tied to distinct contexts \citep{KeysarHayakawaAn2012}. Our goal is to determine whether these trajectories actually differ in practice and if one is superior to others in reasoning outcomes. Similar to how bilingual humans can be instructed to respond in a single language, we can constrain LLMs to generate outputs exclusively in one language. By applying this constraint, we ablate code-switching capabilities from the model, enabling direct comparison between unconstrained bilingual outputs and constrained monolingual outputs in terms of reasoning performance. 

Specifically, during decoding we enforce token-level language constraints by allowing only tokens from the designated language (Fig.~\ref{fig:constrained_decoding}). We apply two types of constraints. In the \textbf{no-switch mode}, we prohibit the model from generating tokens in a particular language by masking those tokens in the vocabulary, which enforces strictly monolingual output. In the \textbf{forced-switch mode}, the model is required to switch languages at a specified token position, at which point only tokens from the target switch-to language are allowed.

\subsection{Constrained vs. Unconstrained Decoding}
\paragraph{Language mixing can enhance reasoning.}
Under the default unconstrained decoding, overall accuracy for \texttt{MATH500} in English and Chinese is balanced. We then compared unconstrained bilingual outputs with constrained monolingual outputs under Chinese prompts (Figure~\ref{fig:bar1}(a)) and English prompts (Figure~\ref{fig:bar1}(b)). Forcing responses to be monolingual Chinese under Chinese prompts reduces accuracy by 5.6 pp. A paired t‑test confirms this drop is statistically significant (p = 0.0017). These results provide strong evidence that LLMs can reason more effectively with language mixing than when restricted to monolingual outputs on certain reasoning tasks, potentially by leveraging the strengths of both languages.

\paragraph{Language mixing may also hurt reasoning.} We then analyzed responses on \texttt{Gaokao Cloze} problems (Figure~\ref{fig:bar2}). Unconstrained responses to Chinese prompts outperform those to English prompts. This is as expected, since Gaokao-like problems (from the Chinese college entrance examination) would predominantly appear in Chinese within the pretraining data.

But contrary to our observations in \texttt{MATH500}, constrained monolingual Chinese decoding outperforms unconstrained bilingual decoding (Figure~\ref{fig:bar2}(a)). We attribute this to an imbalance in monolingual reasoning capabilities where Chinese performance exceeds English for these problems. However, the model still defaults to switching into English, a strategy that is generally helpful but detrimental in this setting.

These negative results for \texttt{Gaokao Cloze} do not imply that language mixing is inherently harmful. Rather, they suggest that QwQ32B‑preview’s built‑in mixing strategy is suboptimal. In the next section, we show that \textit{strategic} language mixing, guided by a probe to decide when to switch, consistently improves LLM performance across datasets.

\section{Can we steer the model toward strategic language mixing?}\label{sec:probe}
\subsection{Probe-Guided Decoding}
\label{sec:probe-codeswitch}

As we’ve shown in the previous section, language mixing is not always beneficial for reasoning. Code-switching can help, harm, or have no impact on the reasoning trajectories, which consequently impacts the overall reasoning outcome. Harmful code-switches may disrupt coherent reasoning chains, while helpful ones can reduce cognitive demands, address lexical gaps, or beneficially reset problematic reasoning directions. Here, we hypothesize that the beneficial, harmful, or neutral impact of each code-switch follows predictable patterns that could be decoded from model activations during generation.

To quantify the impact of code-switching, we compare full generations with and without a switch at each token position and label switches as \{\texttt{\textcolor{darkgreen}{Beneficial}}, \texttt{Neutral}, or \texttt{\textcolor{red}{Harmful}}\}. We apply constraints at a single token position, either by preventing a natural switch (no-switch mode) or forcing a switch where one would not naturally occur (forced-switch mode). A switch is labeled \texttt{\textcolor{darkgreen}{Beneficial}} if it leads to a correct answer that the monolingual version does not; \texttt{\textcolor{red}{Harmful}} if it causes an otherwise correct answer to become incorrect; and \texttt{Neutral} if it has no effect on the final output.

\begin{figure}[!htbp]  
    \centering
    \includegraphics[width=0.48\textwidth]{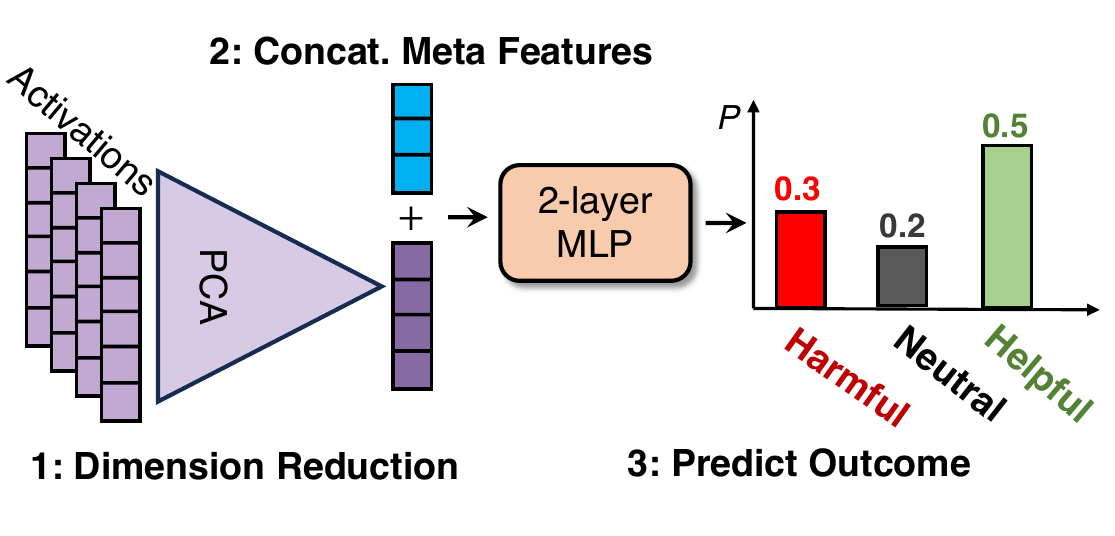}  
    \caption{Architecture of the probe. The model classifies each code-switch as \{\texttt{\textcolor{darkgreen}{Beneficial}}, \texttt{Neutral}, \texttt{\textcolor{red}{Harmful}}\} using hidden activations at the switching step along with meta features.}
    \vspace{-1em}
    \label{fig:probe_architecture}  
\end{figure}

In practice, we collect all natural switching positions and synthesize additional switches at high language entropy positions. We then train a lightweight three-layer MLP probe (Fig.\ref{fig:probe_architecture}) on hidden activations extracted from the LLM. We augment activations with three \textit{meta features}: \ding{182}~\texttt{is\_natural} (natural or synthetic switch), \ding{183}~\texttt{switch\_direction} (Chinese to English or vice versa), and \ding{184}~\texttt{language\_entropy} (entropy of the model’s predicted language distribution).


With the trained probe, we can control the decoding process by predicting online whether a code-switch is beneficial or harmful, and applying token-level constraints accordingly. If a natural switch is classified as \texttt{\textcolor{red}{Harmful}}, we suppress it using constrained decoding in no-switch mode. If a high language entropy position is classified as \texttt{\textcolor{darkgreen}{Beneficial}}, we trigger a forced switch at that step using forced-switch mode. This allows us to steer the model toward strategic language mixing with minimal computational overhead, using only a lightweight and easily deployable MLP probe during decoding.

\subsection{Performance of Probe-Guided Decoding}
\label{sec:probe-training-data}

\paragraph{Probe achieves positive utility score.} 

Since the probe is ultimately used to guide decoding decisions rather than to precisely classify code-switching impacts, its effectiveness should be assessed in terms of its practical impact on multilingual reasoning, i.e., the overall performance gain compared to unconstrained decoding, rather than classification metrics like F1 score. Here, we use decision utility as a measure to approximate the effectiveness of the probe given the helpful/ neutral/ harmful ground truth labels we collected from a large number of reasoning chains with code-switch interventions. (see Appendix~\ref{sec:appendix_utility}).

Specifically, we assign a utility score of $+1$ when the probe correctly classifies a \texttt{\textcolor{red}{Harmful}} or \texttt{\textcolor{darkgreen}{Beneficial}} switch, and a score of $-1$ when it suppresses a \texttt{\textcolor{darkgreen}{Beneficial}} switch or allows a \texttt{\textcolor{red}{Harmful}} one. We achieve positive utility score for all trained datasets, with the highest utility score being $s = 0.0107\approx1/93$. While this score may appear small, each Chinese prompt naturally results in about 8 potential code-switches on average. This means that 1 in 12 questions is expected to benefit directly from a correctly identified helpful switch, corresponding to a potential 8.3 pp gain in accuracy.

\paragraph{Probe-Guided decoding further improves reasoning.} 
We integrated the trained probe with optimal thresholds into our end-to-end decoding pipeline to assess its practical impact on LLM reasoning accuracy. We evaluate this intervention on the \texttt{MATH500} and \texttt{Gaokao Cloze}, where it yields accuracy improvements of 1.62 pp and 2.92 pp, respectively.

We evaluate the probe's effectiveness both within and across datasets. Given the relatively small problem sets, we conducted five random train/test splits to ensure robust evaluation. As shown in Table~\ref{tab:within}, our results demonstrate consistent improvements in both \texttt{MATH500} and \texttt{Gaokao Cloze}. We further applied probes trained on one dataset to different test sets. Results show consistent accuracy improvements across most scenarios (Table~\ref{tab:cross}), demonstrating that our probe learns generalizable patterns that enhance reasoning through strategic language mixing.

\begin{table*}[t]
\caption{Within-dataset evaluation of probe-guided decoding on \texttt{MATH500} and \texttt{Gaokao Cloze}. 
Accuracy gain is reported in percentage points (pp).}
\centering
\small
\renewcommand{\arraystretch}{1.1}
\begin{tabular}{lcc}
\toprule
\textbf{Dataset} & \textbf{Test Utility (mean ± std)} & \textbf{Accuracy Gain (pp; mean ± std)} \\
\midrule
MATH500        & $0.0056 \pm 0.0029$              & $+1.62 \pm 1.01$             \\
Gaokao\_Cloze  & $0.00014 \pm 0.0040$             & $+2.92 \pm 2.12$             \\
\bottomrule
\end{tabular}
\label{tab:within}
\end{table*}

\begin{table*}[ht]
\centering
\small
\caption{Cross-dataset generalization of probe-guided decoding, indicating some transferability across datasets.}
\label{tab:cross}

\begin{tabular}{@{} ccc @{}}
\toprule
\multicolumn{1}{@{}c}{\textbf{Train} $\boldsymbol{\to}$ \textbf{Test}} & 
\multicolumn{1}{c}{\textbf{Test Utility}} & 
\multicolumn{1}{c}{\textbf{Accuracy Gain (pp)}} \\ 
\midrule
MATH500 $\to$ Gaokao\_Cloze            & $-0.0004$      & $+2.12$ \\
Gaokao\_Cloze $\to$ MATH500            & $+0.0024$      & $+3.00$ \\ 
\midrule
\multicolumn{3}{c}{\textit{Evaluation on AIME2024}} \\ 
\midrule
MATH500 $\to$ AIME2024                 & $+0.0036$      & $+0.00$ \\
Gaokao\_Cloze $\to$ AIME2024           & $+0.0054$      & $+3.33$ \\
MATH500 + Gaokao\_Cloze $\to$ AIME2024 & $+0.0033$      & $+3.00$ \\
\bottomrule
\end{tabular}
\end{table*}

\begin{CJK*}{UTF8}{gbsn}

Without improving the LLM's inherent reasoning ability in either language, we achieve noticeable performance improvements solely by training a language decision module (the probe) to guide strategic language mixing. This is analogous to teaching a bilingual speaker to mix languages wisely without teaching them more math—the underlying knowledge remains unchanged, but the strategic mix of languages enhances problem-solving effectiveness. Importantly, these gains are not due to simple language dominance effects. For Gaokao Cloze, where Chinese dominates, our probe-guided strategy does not simply constrain all outputs to Chinese but instead introduces strategic English mixing, with 66.8\% of interventions promoting English usage.

Examining the probe's switching strategy reveals interesting patterns. Examples of helpful switches include converting "柯西-施瓦茨不等式" to "Cauchy-Schwarz inequality" to use more grounded English terminology, and in response to a Chinese prompt, switching from "表达式" to "notation" to create stronger referential coherence with the previously introduced notation (see full context in Appendix~\ref{sec:appendix_probe}).
\end{CJK*}

Overall, as shown in the left panels of Figures~\ref{fig:bar1} and~\ref{fig:bar2}, probe‑guided \textit{strategic} language mixing consistently outperforms the monolingual baseline (dashed line) by selecting near‑optimal mixing strategies that boost reasoning performance.

\section{Related Work}

\paragraph{Multilingual Reasoning in LLMs.} 

As LLMs have evolved from primarily English-centric systems to incorporate more balanced multilingual corpora, they have developed substantial multilingual capabilities~\cite{cui2023efficient, faysse2024croissantllm, yang2024qwen2technicalreport, liu2024deepseek}. However, these models still underperform when reasoning in non-English languages, particularly low-resource ones. This is evidenced by their superior performance on English-translated questions~\cite{shi2022language} and their tendency to switch to English against instructions~\cite{marchisio2024understanding, hinck2024llava, guo2025deepseek}, a limitation long attributed to training data imbalance~\cite{kew2023turning, papadimitriou2022multilingual}. Mechanistic interpretability studies have investigated whether multilingual LLMs truly reason in non-English languages, revealing that some models can "think" in latent non-English languages for specific tasks~\cite{wendler2024llamas, zhong2024beyond} and that distinct language-specific neural circuits exist within these systems~\cite{zhao2024large, tang2024language, zhang2024same}. With the same aim of understanding multilingual reasoning in LLMs, we focus on bilingual code-switching \textit{within} responses to examine how polyglot models reason differently from proficient monolingual systems.

\paragraph{Code-Switching in LLMs.}
Code-switching can emerge in LLMs from exposure to human-generated mixed-language text in training corpora~\cite{wang2025investigating}. The unintended mixing of languages in LLM outputs has been negatively characterized as language confusion, primarily observed when models processing low-resource languages shifted toward English during generation~\cite{marchisio2024understanding, chen2024large,wang2025investigating}. 

Only recently have models begun to more frequently mix English and Chinese-two high-resource and structurally distinct languages-within their reasoning chains. This behavior has emerged in models trained with reinforcement learning \cite{guo2025deepseek, qwen2024qwqpreview, xie2025logic}, where optimizing for outcome-based rewards appears to override the preference for monolingual output. Notably, enforcing language consistency in DeepSeek-R1 resulted in a measurable drop in performance, suggesting a trade-off between language consistency and reasoning ability \cite{guo2025deepseek}. Though a follow-up study using a smaller model claimed language mixing harms reasoning, this conclusion was based on a single logic puzzle dataset and lacks generalizable evidence~\cite{xie2025logic}. Given these conflicting findings, our work aims to systematically evaluate the impact of code-switching on reasoning performance.


\section{Conclusion}
We investigate the impact of English-Chinese language mixing on LLM reasoning. First, we find that language mixing correlates with question difficulty, and by tracing model development, we show that RLVR training is the key trigger for this behavior. To establish causality between increased mixing and reasoning gains, we compare unconstrained bilingual decoding to constrained monolingual decoding on \texttt{MATH500}, finding a statistically significant accuracy boost for the bilingual outputs. Next, we train a lightweight probe to predict the utility of each potential switch and incorporate it into the decoding process across all datasets, resulting in consistent performance improvements over monolingual responses. Altogether, these results suggest that language mixing is not a random artifact of multilingual training but a deliberate strategy that LLMs adopt to improve complex reasoning.

For multilingual LLM reasoning, our findings suggest that language mixing can be beneficial, calling for caution before applying rigid language constraints and motivating future causal analyses. More broadly, we propose that language mixing can extend beyond spoken languages, occurring across modalities~\cite{jiang2022draft, chern2025thinkinggeneratedimages}. We encourage future research to explore these broader forms of language mixing in LLMs.
\section{Limitations}
Our study is limited in scope. The benchmarks we tested are limited to math tasks, and evaluating other domains such as science or logic puzzles is needed to assess the generality of our conclusions on LLM reasoning. We only focus on English-Chinese mixing, and it remains an open question whether similar patterns extend to other language pairs. Intervention analyses focus on QwQ32B-Preview, as broader evaluation across models is limited by the lack of access to RL-trained models that exhibit language mixing (public models such as DeepSeek-R1 and its distilled variants are constrained by enforced language consistency).

We also lack strictly controlled comparisons across training stages. RLVR models are typically trained with additional stages such as SFT, preventing us from isolating the specific contribution of RLVR. Our claim that RLVR promotes language mixing is therefore preliminary, and future work should test this more systematically on controlled model variants.

Finally, our use of hard constrained decoding may inherently reduce performance by imposing an extra language constraint. Future work could explore finer or continuous control over switching frequency and provide stronger empirical comparisons between unconstrained and constrained decoding.

\bibliography{main}
\appendix
\newpage
\section{Appendix}
\subsection{Overall Setup}

All model inference is conducted using half-precision (float16) quantization. For the Qwen model series, we run inference on either three NVIDIA V100 GPUs (32 GB each) or a single NVIDIA A100 GPU (80 GB). Tasks involving probing and probe-guided decoding, which require extracting model activations are executed exclusively on the A100 GPUs. The DeepSeek family of models (including V3 and R1 variants) are queries with OpenRouter API.

Decoding is performed using greedy decoding with a temperature of 0.0, ensuring deterministic outputs across runs. We use flexible token limits, allowing each response to continue until the model emits an end-of-sequence token. In practice, we iteratively doubled the token budget until EOS or until a looping pattern was detected. 

\subsection{Evaluation datasets}
\begin{table*}[ht]
\centering
\caption{Pass@1 and token usage statistics by dataset and prompt language.}
\label{tab:pass1_token_stats}
\begin{tabular}{@{}l l c l@{}}
\toprule
\textbf{Dataset} & \textbf{Prompt Language} & \textbf{Pass@1} & \textbf{Token Count (mean $\pm$ std, max)} \\
\midrule
\multirow{2}{*}{\texttt{MATH500}}      & Chinese & 0.906 & $2093.32 \pm 3492.81$ (32768) \\
                                        & English & 0.926 & $2224.41 \pm 2609.91$ (32768) \\
\addlinespace
\multirow{2}{*}{\texttt{AIME2024}}     & Chinese & 0.367 & $11094.57 \pm 10110.03$ (32768) \\
                                        & English & 0.533 & $6741.53 \pm 3738.44$ (16384) \\
\addlinespace
\multirow{2}{*}{\texttt{GAOKAO\_CLOZE}} & Chinese & 0.877 & $1516.70 \pm 1050.03$ (5972) \\
                                        & English & 0.843 & $2473.74 \pm 1811.84$ (8192) \\
\bottomrule
\end{tabular}
\label{tab:appendix_performance}
\end{table*}

We evaluate model behavior across four math reasoning benchmarks:
\begin{itemize}
\item{Math500~\cite{lightman2023let}: A curated dataset of 500 high school and early undergraduate-level math word problems, designed to test symbolic reasoning and arithmetic across diverse topics.}
\item{AIME2024~\cite{aime2024} Invitational Mathematics Examination): A benchmark of 30 challenging math problems targeted at advanced high school students.}
\item{Gaokao Cloze}: A set of standardized math questions from the Chinese college entrance examination. These problems are extracted from the AGIEval benchmark~\cite{zhong2023agieval}. Gaokao Cloze contains 118 fill-in-the-blank problems.

\end{itemize}

All problems are translated into both English and Chinese to support code-switching and bilingual evaluation. Translations are first generated using GPT-4o, followed by manual review and correction.

We evaluate model performance using Pass@1 accuracy, defined as the percentage of problems correctly solved on the first attempt. Specifically, We extract the final answer from the model’s output using a pattern-based parser ($\backslash boxed$) and compare it against the ground truth. For outputs that do not exactly match the reference, we perform manual evaluation to account for correct answers expressed in alternative formats. When compared to the QwQ32B-preview~\cite{qwen2024qwqpreview} report (MATH500 90.6\%, AIME 50.0\%), we observe no discrepancies in reasoning outcome (Table~\ref{tab:appendix_performance}).

\subsection{RLVR Triggers Language Mixing}\label{sec:appendix_rlvr}
\paragraph{Identifying training stages that trigger language mixing.}
To ensure a fair comparison, we explicitly prompt both base and SFT/RLHF models to produce extended chain‑of‑thought reasoning. On MATH500, their average token counts closely match those of the RLVR models but remain slightly lower (Table~\ref{tab:appendix_token}).
\definecolor{lightgray}{gray}{0.9}
\definecolor{lightpeach}{RGB}{255,228,225}
\begin{table*}[t]
\centering
\caption{Mean token counts for Chinese (CH) and English (EN) prompts on MATH500 (float16, V100×3, truncated at 4096).}
\label{tab:token-counts}
\begin{tabular}{lrr}
\toprule
\textbf{Model} 
  & \textbf{CH Token Count} 
  & \textbf{EN Token Count} \\
\midrule
\rowcolor{white}
\textbf{Qwen2.5‑32B (Length-matched)}
  & 1041.37   &  970.75   \\

\rowcolor{lightgray}
\textbf{Qwen2.5‑32B‑Instruct (Length-matched)}
  &  671.66   & 1008.99   \\

\rowcolor{lightpeach}
QwQ32B‑Preview
  & 1580.44   & 1869.58   \\

\rowcolor{lightpeach}
QwQ32B
  & 3419.92   & 3546.39   \\
\midrule
\rowcolor{white}
\textbf{DeepSeek‑V3‑Base (Length-matched)}
  & 1556.07   & 1792.66   \\

\rowcolor{lightgray}
\textbf{DeepSeek‑V3 (Length-matched)}
  & 1441.09   & 1604.09   \\

\rowcolor{lightpeach}
DeepSeek‑R1‑Zero
  & 1863.33   & 2145.74   \\

\rowcolor{lightpeach}
DeepSeek‑R1
  & 2413.32   & 2457.40   \\

\rowcolor{lightpeach}
DeepSeek‑R1‑Distill‑Llama‑8B
  & 2467.37   & 2600.42   \\

\rowcolor{lightpeach}
DeepSeek‑R1‑Distill‑Qwen‑32B
  & 2185.93   & 2473.86   \\
\bottomrule
\end{tabular}
\label{tab:appendix_token}
\end{table*}

\paragraph{How does RLVR trigger language mixing?} 
We examine RLVR checkpoints at successive training steps, where outcome rewards rise monotonically. Using Tina‑Open‑RS1 (LoRA+GRPO on DeepSeek‑R1‑Distill‑Qwen‑1.5B)~\cite{wang2025tinatinyreasoningmodels}, we generate responses to a held‑out set of English MATH500 problems and compare in Table~\ref{tab:appendix_rlvr}(a) degree of language mixing and (b) average reward relative to group average for code-switching (CS) vs monolingual (Mono) responses. We find that:
\begin{itemize}
    \item Language mixing increases progressively with RL training steps (25.0\% -> 100\%).
    \item Code-switching responses consistently outperform monolingual responses within trajectory groups (Avg Score CS > Avg Score Mono). GRPO thus upweight these higher-advantage code-switching responses, reinforcing the mixing behavior.
\end{itemize}

\begin{table*}[htbp]
\centering
\caption{Evolution of Code-Switching (CS) and Monolingual (Mono) performance across RL steps. Mean scores relative to the group average are shown in parentheses.}
\label{tab:appendix_rlvr}
\begin{tabular}{@{} cccc @{}} 
\toprule
\textbf{RL Step} & \textbf{Avg. Score (CS)} & \textbf{Avg. Score (Mono)} & \textbf{\% CS} \\ 
\midrule
00    & 0.438 (+0.118) & 0.281 (-0.039) & 25.00\%  \\
400   & 0.524 (+0.047) & 0.386 (-0.091) & 65.63\%  \\
600   & 0.662 (+0.107) & 0.407 (-0.148) & 57.81\%  \\
800   & 0.681 (+0.064) & 0.441 (-0.176) & 73.44\%  \\
1200  & 0.555 (-0.007) & 0.611 (+0.049) & 85.94\%  \\
1800  & 0.688 (+0.000) & N/A            &100.00\%  \\
2400  & 0.688 (+0.000) & N/A            &100.00\%  \\
\bottomrule
\end{tabular}
\end{table*}

\subsection{Rule-Based Code-Switch Detection}\label{sec:appendix_detection}

First, we remove domain-specific mathematical content, which is language-agnostic by nature. We strip LaTeX-style math expressions using regular expressions that match content enclosed in dollar signs, \verb|(|) delimiters, and \verb|[|] environments. In addition, we filter digits, parentheses, brackets, mathematical operators, and Greek letters, as these symbols are typically language-agnostic in reasoning contexts.

Second, we detect language boundaries by identifying continuous runs of characters belonging to either the Chinese Unicode range (\verb|U+4E00| to \verb|U+9FFF|) or ASCII alphabetic characters. To improve precision, we implement the following filtering rules for English token candidates: \ding{182}~We exclude domain-specific terms common in mathematical discourse, including mathematical functions (e.g., $\mathrm{sin}$, $\mathrm{cos}$, $\mathrm{ln}$), standard variable names (e.g., $\mathrm{ab}$, $\mathrm{bc}$), and geometric designations (e.g., $\mathrm{ABCD}$). \ding{183}~Single-letter English tokens are discarded to prevent false positives from isolated variable names. \ding{184}~All-capitalized sequences of 2-3 characters are filtered, as these typically represent geometric entities rather than English words.

Finally, we identify language transitions by tracking adjacent language segments. A code-switch is recorded when the language classification of adjacent valid segments changes (e.g., from Chinese to English or vice versa). For each switch, we capture the switch direction, the text content at the boundary, and the position within the full response. Additionally, we track the starting language of each response to establish the baseline language context.

\subsection{Detailed Implementation of Constrained Decoding}
We apply constrained decoding using the same detection rule as in Appendix~\ref{sec:appendix_detection}, but perform code‐switch detection online. Our state‐transition model is defined as follows:

\begin{itemize}
  \item \textbf{Math mode}: Enter math mode when a left bracket—\texttt{\{} , \texttt{[} or \texttt{(}—is detected. Exit math mode when the matching right bracket—\texttt{\}} , \texttt{]} or \texttt{)}—appears.
  \item \textbf{Chinese/English mode}: Switch to Chinese mode if a token’s Unicode code point falls within the CJK range. Otherwise, remain or switch to English mode.
\end{itemize}

To handle composite Chinese tokens—pairs of tokens that only form Chinese characters when combined—we scan for these specific sequences. Upon detection of any such composite token, we transition into Chinese mode. 

\subsection{Probing for Beneficial Code-Switches}
To identify code-switching positions that are beneficial to reasoning accuracy, we train a lightweight probe (Figure \ref{fig:probe_architecture}) on hidden representations extracted from \texttt{QwQ-32B-Preview}. Specifically, we concatenate activations from a selected set of transformer layers and project it into a lower-dimensional space using a PCA transformation fitted on the training set. We also add three related \textit{Meta Features}: \ding{182} \texttt{is natural} (whether a switch is natural or synthetic), \ding{183}~\texttt{switch direction} (whether the switch direction is from Chinese to English or not),  \ding{184}~\texttt{language entropy} (the entropy calculated from the model probability of output Chinese or English token). These feature are appended to the hidden embedding after PCA.

The probe comprises two separate three‑layer MLP heads—one for “no‑switch” and one for “forced‑switch” decoding. We train it by jointly minimizing (1) the negative decision utility and (2) a weighted cross‑entropy loss against an approximated three‑class distribution (beneficial/neutral/harmful). Rather than using one‑hot labels, we infer this distribution from the change in token count: if decoding without a switch produces substantially more tokens, we treat code‑switching as more likely to be beneficial.

To address class inbalance, we use a class weight of \{$1.0$, $0.1$, $1.0$\} for \{\texttt{\textcolor{darkgreen}{Beneficial}}, \texttt{Neutral}, \texttt{\textcolor{red}{Harmful}}\} to downweight the majority class (\texttt{Neutral}) and upweight the minority class (\texttt{\textcolor{darkgreen}{Beneficial}} and \texttt{\textcolor{red}{Harmful}}). During inference, the probe outputs predicted probabilities for each class after softmax. To maximize decision utility, we apply a thresholding strategy: if the predicted probability of a harmful switch exceeds a threshold $\tau_{\text{harm}}$, we suppress the switch; if the predicted probability of a beneficial switch exceeds a threshold $\tau_{\text{help}}$, we enforce the switch. Thresholds are selected via a grid search on a held-out validation set to maximize a custom utility metric that penalizes missed beneficial switches and incorrectly allowed harmful switches.

\subsection{Probe Performance and Utility}\label{sec:appendix_utility}
\paragraph{Training Data Collection and Statistics.} 
We collect training data for the probe using token-level constrained decoding focused on positions of natural switches or with high language entropy. We begin by identifying natural code-switching positions and apply the \textbf{no switch mode} to collect examples where switching is suppressed. If the number of natural switches falls short of a threshold, we supplement the dataset by introducing synthetic switches using the \textbf{forced switch mode}. We provide detailed statistics of the activation data we collected for \texttt{Math500} and \texttt{Gaokao Cloze} in Table \ref{tab:math500_class_distribution} and Table \ref{tab:gaokao_cloze_class_distribution}, respectively. The statistics reveal a strong class imbalance: the majority of code-switching instances fall into the \texttt{Neutral} category.

\begin{table}[ht]
\centering
\caption{Class distribution of the MATH 500 dataset across train, validation, and test splits.}
\label{tab:math500_class_distribution}
\begin{tabular}{lccc}
\toprule
\textbf{Class}  & \textbf{Train} & \textbf{Validation} & \textbf{Test} \\
\midrule
Harmful & 773 & 127 & 204 \\
Neutral & 7,120 & 803 & 1,699 \\
Helpful & 894 & 125 & 241 \\
\midrule
\textbf{Total} & \textbf{8,787} & \textbf{1,055} & \textbf{2,144} \\
\bottomrule
\end{tabular}
\end{table}


\begin{table}[ht]
\centering
\caption{Class distribution of the Gaokao Cloze dataset across train, validation, and test splits.}
\label{tab:gaokao_cloze_class_distribution}
\begin{tabular}{lccc}
\toprule
\textbf{Class}  & \textbf{Train} & \textbf{Validation} & \textbf{Test} \\
\midrule
Harmful & 172 & 23 & 73 \\
Neutral & 1,427 & 199 & 394 \\
Helpful & 260 & 36 & 86 \\
\midrule
\textbf{Total} & \textbf{1,859} & \textbf{258} & \textbf{553} \\
\bottomrule
\end{tabular}
\end{table}

The language entropy at each token position is defined as, where $p'_en$ and $p'_ch$ is reweighted to sum to 1.
\[
  H \;=\; -\,p'_{\mathrm{en}}\log_{2}\bigl(p'_{\mathrm{en}}\bigr)\;-\;p'_{\mathrm{ch}}\log_{2}\bigl(p'_{\mathrm{ch}}\bigr),
\]
where \(p'_{\mathrm{en}}\) and \(p'_{\mathrm{ch}}\) are renormalized to satisfy \(p'_{\mathrm{en}} + p'_{\mathrm{ch}} = 1\).  
\[
  H_{\mathrm{final}} \;=\; \bigl(p_{\mathrm{en}} + p_{\mathrm{ch}}\bigr)\;H.
\]

This ensures that we only consider positions where both CH and EN are probable and uncertainty is high.

\paragraph{Hyperparameters and Experimental Setup.}
We use a stratified train/validation/test split of $70\%$/$10\%$/$20\%$ by problem ID, ensuring that code-switch examples from the same problem do not appear in multiple splits. All experiments are conducted on a single NVIDIA A100 GPU. The probe uses intermediate layer activation derived from five transformer layers (layers 63, 47, 31, 15, and 0), with additional metadata features. We reduce the input dimensionality using PCA, followed by a projection layer of dimension 512 and a hidden layer of size 512. Training is run for 30 epochs with a batch size of 256 and a learning rate of $1\text{e}{-4}$.

\begin{table}[ht]
\centering
\caption{Hyperparameters for the probe model used across all datasets.}
\label{tab:math500_probe_hyperparams}
\begin{tabular}{lc}
\toprule
\textbf{Hyperparameter} & \textbf{Value} \\
\midrule
Selected Layers & [63, 47, 31, 15, 0] \\
Use Metadata & True \\
PCA Dimension & 512 \\
Projection Dimension & 512 \\
Hidden Dimension & 512 \\
Coefficient for CE loss & 0.5 \\
Class Weights & [1.0, 0.1, 1.0] \\
Number of Epochs & 30 \\
Batch Size & 256 \\
Learning Rate & $1 \times 10^{-4}$ \\
\bottomrule
\end{tabular}
\end{table}

\paragraph{Decision Utility.} 
 We use decision utility as an approximate measure of the probe’s effectiveness, based on our ground-truth labels of helpful, neutral, and harmful interventions collected from extensive reasoning chains with code-switch interventions.


\begin{table*}[ht]
\caption{The utility matrix for Case 1: natural switch (we block when pred = Harmful)}
\centering
\begin{tabular}{c|ccc}
\toprule
GT $\downarrow$\ Pred$\rightarrow$ & \textbf{Harmful (H)} & \textbf{Neutral (N)} & \textbf{Beneficial (B)} \\
\midrule
\textbf{Harmful (H)}    & +1 &  0 &  0 \\
\textbf{Neutral (N)}    &  0 &  0 &  0 \\
\textbf{Beneficial (B)} & -1 &  0 &  0 \\
\bottomrule
\end{tabular}
\label{tab:appendix_mat1}
\end{table*}

\begin{table*}[ht]
\centering
\caption{The utility matrix for Case 2: no natural switch (we inject when pred = Beneficial)}
\begin{tabular}{c|ccc}
\toprule
GT$\downarrow$\ Pred$\rightarrow$ & \textbf{Harmful (H)} & \textbf{Neutral (N)} & \textbf{Beneficial (B)} \\
\midrule
\textbf{Harmful (H)}    &  0 &  0 & -1 \\
\textbf{Neutral (N)}    &  0 &  0 &  0 \\
\textbf{Beneficial (B)} &  0 &  0 & +1 \\
\bottomrule
\end{tabular}
\label{tab:appendix_mat2}
\end{table*}

\begin{figure*}[!htbp]
  \centering
  \begin{minipage}[t]{1\textwidth}
    \centering
    \includegraphics[width=\textwidth]{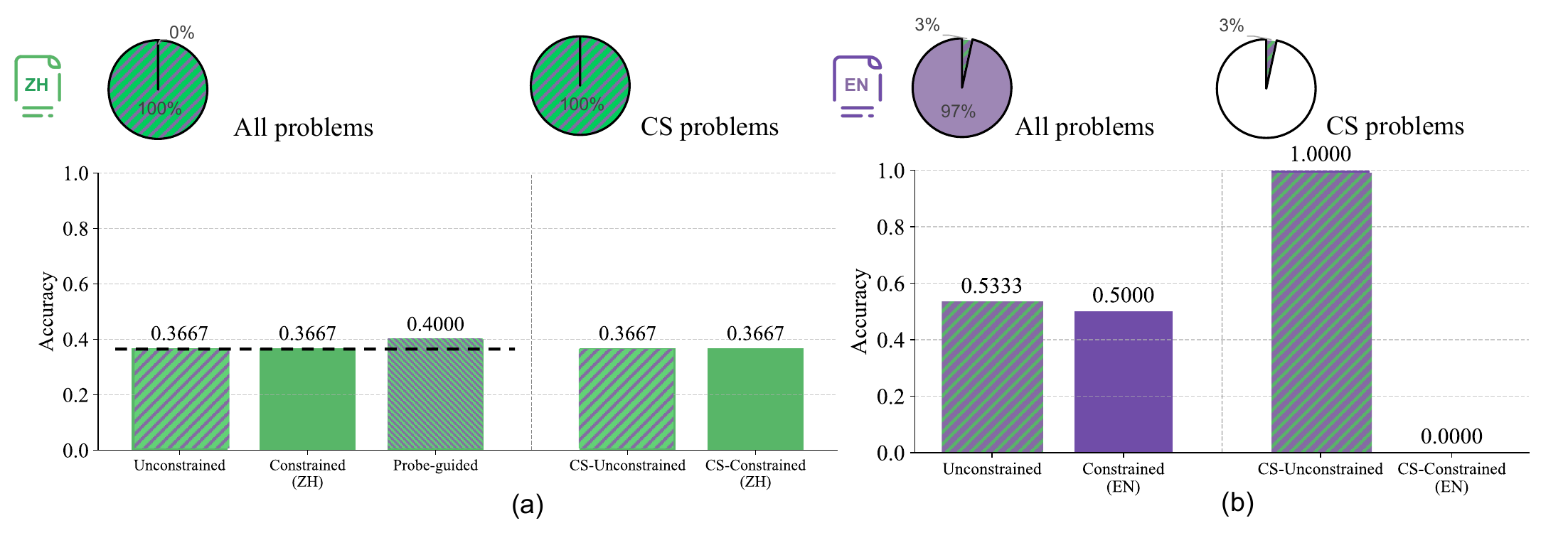}
  \end{minipage}%
  \caption{Comparison of accuracies for unconstrained, constrained, and probe‑guided decoding on \texttt{AIME2024}. \textbf{(a)} Chinese prompts (green); \textbf{(b)} English prompts (purple).}
  \label{fig:bar3}
\end{figure*}
These utility matrices (Table~\ref{tab:appendix_mat1} and~\ref{tab:appendix_mat2}) reveal a strong asymmetry in decision costs—for example, in Case 1 any non‑harmful prediction (N or B) yields zero utility, even though a balanced F1 would penalize N vs B errors equally. By focusing on \emph{decision utility}, we target only those errors that actually affect decoding outcomes, making it a more appropriate measure than balanced classification metrics.

\paragraph{Evaluating Performance Gain.}

The lightweight probe is integrated into the decoding loop at each step. At every token position, we first generate the model’s natural output without intervention, then classify the position into Case 1—where a natural code‐switch occurred—or Case 2—where no switch was generated but the language entropy exceeds a predefined threshold. In Case 1, the probe issues a decision to suppress the detected switch; in Case 2, it decides whether to inject a switch into the decoding stream.

With a strict train/validation/test partition, we measure net gains by contrasting probe‑guided decoding with unconstrained decoding on the test set across five random splits. In Figures~\ref{fig:bar1} and~\ref{fig:bar2}, the “probe‑guided” bars represent the unconstrained baseline plus the average gain over those splits—not the raw performance, which would be inflated by training‑set leakage.

\subsection{Performance Comparison on AIME2024}
We compare reasoning performance across unconstrained decoding, constrained monolingual decoding, and probe‑guided decoding in Figure~\ref{fig:bar3}. For this experiment, the probe is trained solely on \texttt{Gaokao Cloze} and applied to \texttt{AIME2024} with no AIME2024 training data. Unlike Figures~\ref{fig:bar1} and~\ref{fig:bar2}, the probe‑guided bar in Figure~\ref{fig:bar3} shows raw performance, revealing a 3.33 pp net gain. This demonstrates both the probe’s cross‑dataset generalization and the consistent benefit of strategic language mixing.

\subsection{Examples of Language Mixing in QwQ-32B-Preview}\label{sec:appendix_examples}

\begin{CJK*}{UTF8}{gbsn}
\begin{tcolorbox}[colback=gray!10, colframe=gray!50, boxrule=0.5pt, arc=2mm]
\textbf{Question:} 将点 \( (0,3) \) 从直角坐标系转换为极坐标系，需要找到对应的极径 \( r \) 和极角 \( \theta \)。极坐标表示为 \( (r, \theta) \)，其中 \( r > 0 \) 且 \( 0 \le \theta < 2\pi \)。

\textbf{Answer:} 首先，回忆一下直角坐标和极坐标的转换公式：

1. \( r = \sqrt{x^2 + y^2} \)

2. \( \theta = \tan^{-1}\left(\frac{y}{x}\right) \)

但是，这些公式在某些情况下需要特别处理，比如当 \( x = 0 \) 时，因为除法会 \textcolor{red}{undefined}。

\textcolor{red}{given point is (0,3), so \( x = 0 \) and \( y = 3 \)}。

首先计算 \( r \):

\( r = \sqrt{0^2 + 3^2} = \sqrt{9} = 3 \)

所以，\( r = 3 \)

接下来是 \( \theta \):

通常，\( \theta = \tan^{-1}\left(\frac{y}{x}\right) \)，但在这里 \( x = 0 \)，所以这个表达式 undefined。

这意味着点在 y 轴上。具体来说， \textcolor{red}{since \( x = 0 \) and \( y = 3 \)， which is positive y-axis}。

在极坐标系中，θ 是从正 x 轴顺时针或逆时针测量到径向线的角度。

对于 \textcolor{red}{positive y-axis, θ is \( \frac{\pi}{2} \) radians (or 90 degrees)}。

所以，\( \theta = \frac{\pi}{2} \)

因此，极坐标是 \( \left(3, \frac{\pi}{2}\right) \)

但为了确保没有错误，让我再检查一下。

点 (0,3) 位于直角坐标系的 \textcolor{red}{positive y-axis }上。

在极坐标中，r 是点到原点的距离，\textcolor{red}{which is 3}, 正确。

θ 是从 \textcolor{red}{positive x-axis} 到连接原点和点的线段的角度，顺时针或逆时针，取决于约定。在这里，我们使用标准的数学约定，即逆时针方向。

从 \textcolor{red}{positive x-axis }逆时针转到 \textcolor{red}{positive y-axis} 是 \( \frac{\pi}{2} \) radians。

所以，\( \theta = \frac{\pi}{2} \) 是正确的。

另外，需要确认 θ 的范围是 \( 0 \le \theta < 2\pi \)，\textcolor{red}{which \( \frac{\pi}{2} \) satisfies}。

所以，最终答案是 \( \left(3, \frac{\pi}{2}\right) \)

\textcolor{red}{**Final Answer**}

\[ \boxed{\left(3,\,\frac{\pi}{2}\right)} \]

\end{tcolorbox}
\end{CJK*}
\begin{CJK*}{UTF8}{gbsn}
\begin{tcolorbox}[colback=gray!10, colframe=gray!50, boxrule=0.5pt, arc=2mm]
\textbf{Question}: In the land of Ink, the money system is unique. One Trinket is equal to 4 Blinkets, and 3 Blinkets are equal to 7 Drinkets. In Trinkets, what is the value of 56 Drinkets?

\textbf{Answer}: Let's tackle this problem step by step. So, in the land of Ink, they have these different units of money: Trinkets, Blinkets, and Drinkets. The relationships between them are given, and I need to find out what 56 Drinkets are worth in Trinkets.

First, I need to understand the relationships between these units. It says that one Trinket is equal to 4 Blinkets. So, if I have one Trinket, it's the same as having 4 Blinkets. Got it.

Next, it says that 3 Blinkets are equal to 7 Drinkets. So, 3 Blinkets = 7 Drinkets. That means that one Blinkey must be worth 7/3 Drinkets, but maybe I don't need to go that far yet.

I need to find out what 56 Drinkets are worth in Trinkets. To do that, I probably need to convert Drinkets to Blinkets first, and then convert Blinkets to Trinkets.

So, let's start by converting Drinkets to Blinkets. If 3 Blinkets = 7 Drinkets, then how many Blinkets are there in 56 Drinkets?

I can set up a proportion for this. If 3 Blinkets are equal to 7 Drinkets, then x Blinkets are equal to 56 Drinkets. So, 3/7 = x/56.

Wait, actually, that might not be the right way to set it up. If 3 Blinkets = 7 Drinkets, then the ratio of Blinkets to Drinkets is 3:7. So, to find out how many Blinkets correspond to 56 Drinkets, I can set up the proportion like this:

...

Then, to find out how many Trinkets are in 56 Drinkets, I divide 56 by the number of Drinkets per Trinket, which is 28/3.

So, Trinkets = 56 / (28/3) = 56 * 3/28 = 6 Trinkets.

\textcolor{red}{一致的结果是}6 Trinkets.

I think it's safe to conclude that 56 Drinkets are worth 6 Trinkets.

**Final Answer**

\[ \boxed{6} \]

\end{tcolorbox}
\end{CJK*}
\subsection{Examples of Probe-Guided Decoding}\label{sec:appendix_probe}

\begin{CJK*}{UTF8}{gbsn}
In the example below, the model initially switches to English for “Wait, no, in standard notation…,” then immediately switches back to Chinese with “所以，根据标准 notation.”\textit{(So, according to the standard notation.)} The probe intervenes at “表达式,” forcing it into English so that “notation” (marked in red) clearly refers back to the earlier “notation” mentioned in the preceding sentence.

\begin{tcolorbox}[colback=gray!10, colframe=gray!50, boxrule=0.5pt, arc=2mm]
\textbf{Question:} 在 $\triangle ABC$ 中, $B=60^{\circ}, AC=\sqrt{3}$, 则 $AB+2BC$ 的最大值为 $(\quad)$.

\textbf{Answer:}
在三角形ABC中，已知角B=60度，边AC=√3，要求AB + 2BC的最大值。

首先，这是个三角形的问题，已知一个角和对边，可以考虑用正弦定理或者余弦定理来处理。

已知角B和边AC（即边a，因为对角B），根据正弦定理：

a / sinB = b / sinA = c / sinC = 2R 

其中，a=√3，B=60°，R是外接圆半径。

所以，2R = a / sinB = √3 / sin60° = √3 / (�√3/2) = 2

所以，b = 2R sinA = 2 sinA

c = 2R sinC = 2 sinC

所以，AB = c = 2 sinC

BC = a = √3

\textbf{Wait, no, in standard notation}:

a = BC, b = AC, c = AB

所以，根据标准 \textcolor{red}{notation}：

b = AC = √3

a = BC

c = AB

所以，a / sinA = b / sinB = c / sinC = 2R

所以，2R = b / sinB = √3 / sin60° = √3 / (√3/2) = 2

...

**最终答案**\[\boxed{2\sqrt{7}}\]

\end{tcolorbox}
In the following example, the probe switches “优雅” to “elegant” in the sentence "\textit{I prefer to find a more elegant approach.}” In Chinese, calling a solution “elegant” can sound immodest, but English freely uses “elegant” in technical contexts. By switching languages, the probe avoids cultural awkwardness and uses a term that feels natural for describing a mathematical solution.

\begin{tcolorbox}[colback=gray!10, colframe=gray!50, boxrule=0.5pt, arc=2mm]
\textbf{Question:}
对于某些$x$值，$0 < x < 180\text{，}$ 

\[
\tan 53^\circ \tan 81^\circ \tan x^\circ
\]
\[
= \tan 53^\circ + \tan 81^\circ + \tan x^\circ.
\]

求$x$。

\textbf{Answer:}

我遇到了一个三角函数的问题，题目是：

对于某些x值，0 < x < 180，有

tan 53° * tan 81° * tan x° = tan 53° + tan 81° + tan x°。

求x的值。

首先，我需要理解这个等式的意思。看起来是一个关于正切函数的等式，涉及到三个角度：53度、81度和x度。这些角度的正切值之间有某种乘积等于它们的和。

我回忆了一下三角函数的 identities，看看有没有什么公式可以直接应用在这里。我记得有一个公式是关于三个角的正切加法公式，特别是在当三个角加起来等于180度的时候。… 但是，我更希望找到一个更 \textcolor{red}{elegant} 的方法，而不必计算具体的 tangent values。

Final Answer**\[ \boxed{46^\circ} \]"

\end{tcolorbox}
\end{CJK*}

\end{document}